\documentclass[journal]{IEEEtran}
\usepackage{amsmath,amsfonts}
\usepackage{algorithmic}
\usepackage{algorithm}
\usepackage{array}
\usepackage[caption=false,font=normalsize,labelfont=sf,textfont=sf]{subfig}
\usepackage{textcomp}
\usepackage{stfloats}
\usepackage{url}
\usepackage{verbatim}
\usepackage{graphicx}
\usepackage{cite}

\usepackage{bbm}
\usepackage{tabularx}
\usepackage{booktabs}
\usepackage{soul}
\usepackage[table]{xcolor}
\usepackage{multirow}
\usepackage{pifont} %
\usepackage{hyperref}
\newcommand{\mc}[3]{\multicolumn{#1}{#2}{#3}}
\newcommand{\mr}[3]{\multirow{#1}{#2}{#3}}
\newcommand{\underl}[1]{\underline{#1}}

\begin{document}

\title{Associate Everything Detected: Facilitating Tracking-by-Detection to the Unknown}

\author{Zimeng Fang, Chao Liang, Xue Zhou, Shuyuan Zhu, \IEEEmembership{Member, IEEE}, and Xi Li, \IEEEmembership{Senior Member, IEEE}
\thanks{This work has been submitted to the IEEE for possible publication. Copyright may be transferred without notice, after which this version may no longer be accessible. This work was supported in part by the Natural Science Foundation of China under Grant 62372082, in part by Shenzhen Natural Science Foundation under Grant JCYJ20240813114206010, in part by the Fundamental Research Funds for the Central Universities under Grant ZYGX2024Z017, and in part by Zhejiang Provincial Natural Science Foundation of China under Grant LD24F020016. The associate editor coordinating the review of this article and approving it for publication was Prof. Leyuan Fang. (Zimeng Fang and Chao Liang contributed equally to this work.) (Corresponding authors: Xue Zhou; Xi Li.)}
\thanks{Zimeng Fang and Chao Liang are with the School of Automation Engineering, University of Electronic Science and Technology of China (UESTC), Chengdu 611731, China (e-mail: fangzimeng@std.uestc.edu.cn; 201921060415@std.uestc.edu.cn).}%
\thanks{Xue Zhou is with the Shenzhen Institute of Advanced Study and the School of Automation Engineering, University of Electronic Science and Technology of China (UESTC), Shenzhen 518110, China (e-mail: zhouxue@uestc.edu.cn).}%
\thanks{Shuyuan Zhu is with the School of Information and Communication Engineering, University of Electronic Science and Technology of China (UESTC), Chengdu 611731, China (e-mail: eezsy@uestc.edu.cn).}%
\thanks{Xi Li is with the College of Computer Science and Technology, Zhejiang University, Hangzhou 310027, China (e-mail: xilizju@zju.edu.cn).}%
}


\maketitle

\begin{abstract}
Multi-object tracking (MOT) emerges as a pivotal and highly promising branch in the field of computer vision.
Classical closed-vocabulary MOT (CV-MOT) methods aim to track objects of predefined categories.
Recently, some open-vocabulary MOT (OV-MOT) methods have successfully addressed the problem of tracking unknown categories.
However, we found that the CV-MOT and OV-MOT methods each struggle to excel in the tasks of the other.
In this paper, we present a unified framework, Associate Everything Detected (AED), that simultaneously tackles CV-MOT and OV-MOT by integrating with any off-the-shelf detector and supports unknown categories.
Different from existing tracking-by-detection MOT methods, AED gets rid of prior knowledge (e.g., motion cues) and relies solely on highly robust feature learning to handle complex trajectories in OV-MOT tasks while keeping excellent performance in CV-MOT tasks.
Specifically, we model the association task as a similarity decoding problem and propose a sim-decoder with an association-centric learning mechanism.
The sim-decoder calculates similarities in three aspects: spatial, temporal, and cross-clip.
Subsequently, association-centric learning leverages these threefold similarities to ensure that the extracted features are appropriate for continuous tracking and robust enough to generalize to unknown categories.
Compared with existing powerful OV-MOT and CV-MOT methods, AED achieves superior performance on TAO, SportsMOT, and DanceTrack without any prior knowledge.
Our code is available at \href{https://github.com/balabooooo/AED}{https://github.com/balabooooo/AED}.
\end{abstract}


\section{Introduction}
\label{sec:intro}

Closed-vocabulary multi-object tracking (CV-MOT) methods~\cite{bewley2016simple,wojke2017simple,wang2020towards,liang2022rethinking,sun2020transtrack} are designed to track a predefined, limited set of target classes consistent with training data.
They have achieved remarkable results in CV-MOT datasets~\cite{leal2015motchallenge,milan2016mot16,dendorfer2020mot20,sun2022dancetrack}.
Nevertheless, some practical applications like autonomous driving~\cite{jing2024stt} and augmented reality (AR) require trackers to handle new or unexpected classes.
CV-MOT methods cannot expand the categories during the detection phase.
In contrast, with the help of zero-shot capability of Multimodal Large Language Model (MLLM)~\cite{radford2021learning,regionclip,liu2023grounding}, open-vocabulary multi-object tracking (OV-MOT) methods~\cite{li2023ovtrack,wu2023general,zheng2024nettrack,chu2024zero} can adapt to a broader range of categories, including unseen ones in training.
However, when tracking certain categories, OV-MOT methods are less effective than well-finetuned CV-MOT methods.

To unify CV-MOT and OV-MOT within a single framework, we reformulate the tracking task to associate arbitrary detections under the tracking-by-detection paradigm.
We intend to build our framework based on the "\textbf{A}ssociate \textbf{E}verything \textbf{D}etected" concept and name our method AED which offers plug-and-play compatibility with various off-the-shelf detectors.
AED is trained on certain categories and aims to associate arbitrary detections including unseen categories.
Therefore, a robust association method capable of handling complex matching and long-term ID preservation is crucial for MOT.

According to whether to use appearance features during association, existing tracking-by-detection MOT methods can be classified into two main categories: appearance-independent and appearance-based.

\begin{figure}[!t]
  \centering
  \includegraphics[width=8.8cm]{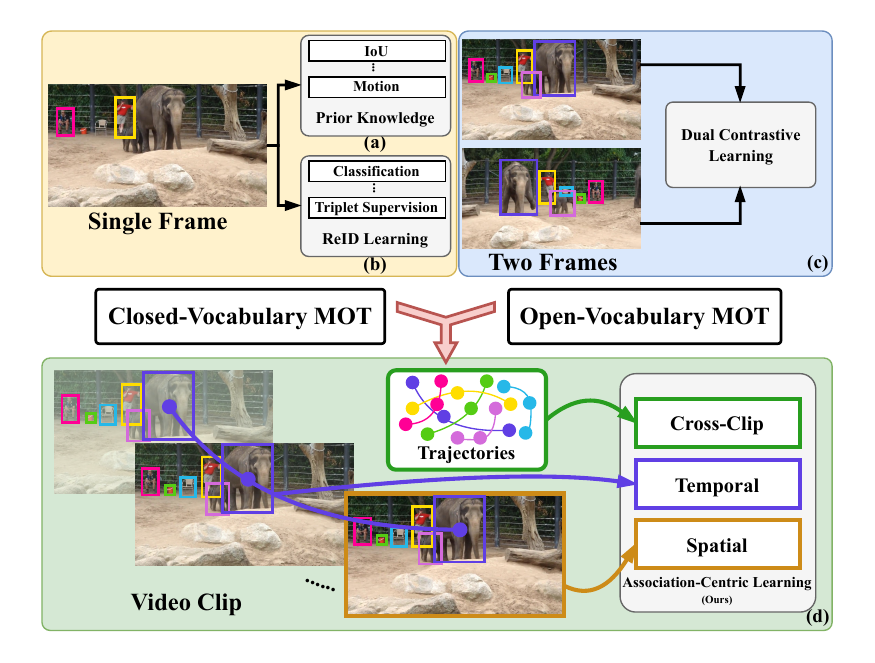}
  \caption{Motivation of AED. Existing closed-vocabulary and open-vocabulary trackers focus on tracking certain categories (e.g., person) and arbitrary categories (e.g., person, elephant, chair, bucket, etc) respectively. For the association phase, (a) appearance-independent methods rely on prior knowledge; (b) some appearance-based CV-MOT methods train their appearance branch following the ReID pipeline; (c) some other appearance-based methods use dual contrastive strategy for feature learning; (d) our proposed method unifies CV-MOT and OV-MOT tasks and uses association-centric learning.
  }
  \label{fig:motivation}
  \vspace{-0.5cm}
\end{figure}

1) Appearance-independent methods~\cite{bewley2016simple,zhang2022bytetrack,cao2023observation} fully utilize prior knowledge to achieve state-of-the-art performance in CV-MOT scenarios~\cite{milan2016mot16,dendorfer2020mot20,sun2022dancetrack} (Fig.~\ref{fig:motivation}(a)).
For example, with high intersection over union (IoU) of objects between adjacent frames, most of these methods operate under the implicit assumption that objects have minimal movement between the two frames.
Additionally, some of them also rely on motion or other prior knowledge, e.g., linear movement~\cite{kalman1960new} and consistent height and confidence variation of objects~\cite{yang2024hybrid}.
However, the reliance on prior knowledge is disastrous in the OV-MOT setting where the motion patterns of different categories exhibit considerable variability.
It is difficult to model these complex motion patterns by using prior knowledge.

2) Some appearance-based methods~\cite{wojke2017simple,liang2022rethinking,seidenschwarz2023simple} leverage the learned re-identification (ReID) features to compensate for the drawback of appearance-independent methods in CV-MOT (Fig.~\ref{fig:motivation}(b)).
In particular, besides using prior knowledge, they train their appearance models or branches following the ReID learning pipeline (e.g., classification or triplet supervision~\cite{hermans2017defense}).
Other methods~\cite{pang2021quasi,li2023ovtrack,wu2023general} (Fig.~\ref{fig:motivation}(c)),  especially for OV-MOT methods, use dual contrastive learning strategy for feature learning between two frames.
However, the learning objectives of both ReID-based and dual contrastive-based methods are inconsistent with the inference objectives of the association stage for MOT.
The association in MOT requires temporal discriminability, while ReID-based methods enable the model to have highly global identity (ID) discriminability by unsuitably modeling the association problem as a classification or alignment task within feature space.
Dual contrastive-based methods lack temporal diversity, i.e., the association stage of MOT usually needs to handle intricate associations across far more than two frames.
We argue that ignorance or lack of association-driven learning may lead to ambiguity during training.

Based on the above analysis, we need to address two key issues to accommodate both CV-MOT and OV-MOT.
1) \textbf{Prior Knowledge Reliance}:
To minimize the reliance on prior knowledge and simplify the tracking logic, we only use appearance cues and model the association task in MOT as the similarity decoding problem.
Specifically, we design a similarity decoder (sim-decoder) where we model the objects in the current frame as object queries and historical trajectories as track queries.
Then we use the sim-decoder to decode the similarities between object queries and track queries.
Without any prior knowledge, the similarity is then used for association directly.
Due to the sufficient reliability of the sim-decoder, we do not employ any complex trajectory initialization strategies during association.
2) \textbf{Association-Learning Ignorance}:
To solve this problem, we propose an association-centric learning mechanism, along with the sim-decoder, to explore training data in three aspects: spatial, temporal, and cross-clip, as illustrated in Fig.~\ref{fig:motivation}(d).
Particularly, contrastive learning is utilized for all three of the aspects above.
Spatial contrastive learning distinguishes IDs within a single frame to lay the groundwork for temporal association.
Temporal contrastive learning matches current frame detections to historical trajectories to exhibit better temporal ID consistency.
Furthermore, cross-clip contrastive learning enhances long-term ID consistency among different short trajectories.

We summarize our contributions as follows:
\begin{itemize}
    \item We propose a unified association-driven tracking-by-detection MOT framework to associate every detected object, bridging the gap between CV-MOT and OV-MOT.
    \item We design a sim-decoder followed by an association-centric learning mechanism to decode highly robust similarities and ensure that the learned features are more appropriate for tracking.
    \item Extensive experiments demonstrate that our method effectively improves the performance in both OV-MOT and CV-MOT tasks, especially in scenarios involving large movements, severe occlusions, similar appearances, and low frame rates.
\end{itemize}

\begin{figure*}[!t]
  \centering
  \includegraphics[height=8cm]{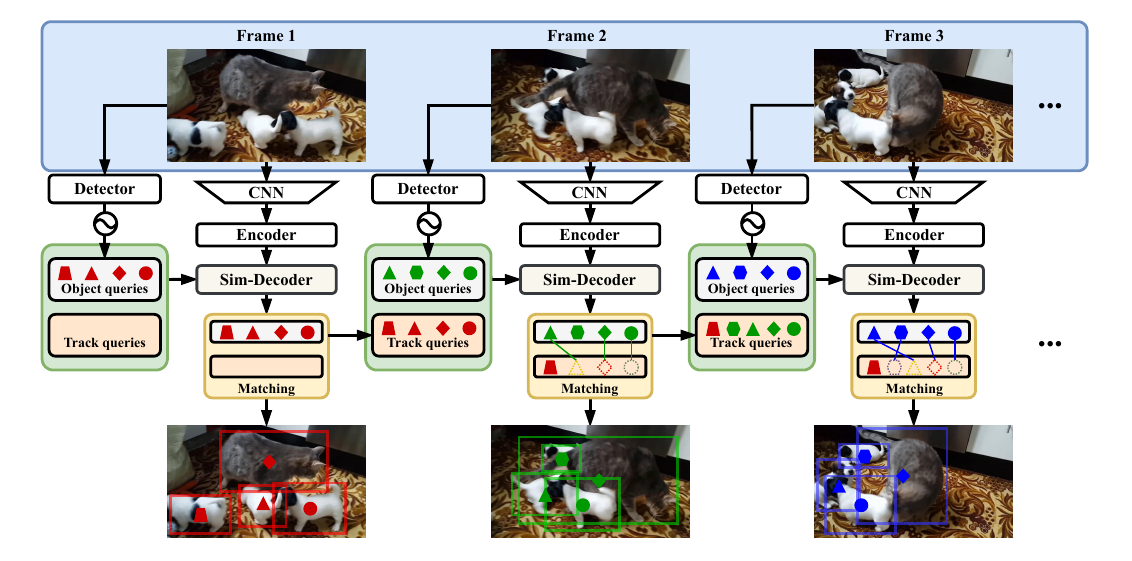}
  \caption{Overview of AED. During inference, a replaceable detector outputs boxes of arbitrary categories (e.g., cat and dog), and boxes are encoded into object queries. A CNN and a transformer encoder are responsible for extracting features of the frame. The sim-decoder calculates the similarities between object queries and track queries. Similarities are then used for matching. If a track query is matched (dashed lines), it will be updated by its object query. Matching results will be the track queries for the next frame. We use different shapes to represent different IDs and different colors within each shape represent frames from different times.
  }
  \label{fig:overview}
  \vspace{-0.5cm}
\end{figure*}

\section{Related Works}
\label{sec:relatedWorks}
In this section, we first briefly review the two most prevailing MOT paradigms, tracking-by-detection and tracking-by-query.
Then we review some related works of Open-Vocabulary MOT.

\subsection{Tracking-by-Detection MOT}
In the past few years, tracking-by-detection has been a popular and high-performance paradigm for CV-MOT tasks.
Objects are first detected and then associated.
Guided by this, many methods have introduced innovative approaches in various aspects, such as motion features~\cite{bewley2016simple,zhang2022bytetrack,cao2023observation,yang2024hybrid,huang2024iterative}, appearance features~\cite{wojke2017simple,seidenschwarz2023simple}, and the real-time performance~\cite{wang2020towards,liang2022rethinking} of the algorithms.
SORT~\cite{bewley2016simple} introduces a Kalman filter~\cite{kalman1960new} to predict the most probable location of objects in the next frame, which effectively solves the association problem.
Based on SORT, DeepSORT~\cite{wojke2017simple} uses an extra re-identification (ReID) model to compensate for the shortcomings of IoU (Intersection over Union) distance.
JDE~\cite{wang2020towards} modifies Yolov3~\cite{redmon2018yolov3} to simultaneously output detection results and ReID features.
CSTrack~\cite{liang2022rethinking} finds the competition between detection and ReID and further proposes a reciprocal network to learn task-dependent representations.
ByteTrack~\cite{zhang2022bytetrack} uses a simple but effective strategy to recover the true positive boxes with low confidence scores.
However, these tracking-by-detection algorithms heavily rely on prior knowledge, especially motion cues (e.g., IoU distance).
Some other advanced methods~\cite{cao2023observation,yang2024hybrid,seidenschwarz2023simple,huang2024iterative} design handcrafted tracking logic (e.g., observation re-update and iterative scale-up IoU) to adapt to specific tracking scenarios such as dancing~\cite{sun2022dancetrack}, pedestrian walking~\cite{milan2016mot16,dendorfer2020mot20}, and sports~\cite{cui2023sportsmot}.
These handcrafted tracking logics are not conducive to open-vocabulary tracking because motion patterns are hard to describe in scenarios with numerous categories.
The reliance on handcrafted tracking logic significantly harms their universality in diverse scenarios.

\subsection{Tracking-by-Query MOT}
Recently, transformer~\cite{vaswani2017attention} has shown its exceptional performance on a wide range of tasks.
Queries of the transformer can aggregate specific features according to different preferences.
Based on this, some existing MOT methods~\cite{sun2020transtrack,zhou2022global,liu2023collaborative} model a trajectory as a query or a set of queries.
Some other tracking-by-query methods~\cite{cai2022memot,gao2023memotr} use the attention mechanism~\cite{vaswani2017attention} to aggregate temporal information.
Then they use queries to detect and track targets across frames end-to-end.
MOTR~\cite{zeng2022motr} advances Deformable DETR~\cite{zhu2020deformable} into a query-based MOT tracker.
MOTRv2~\cite{zhang2023motrv2} improves the detection performance of MOTR by introducing an additional detector.
These trackers are highly integrated.
However, such advantages can also be considered as their shortcomings somehow.
Their parameters, including queries, are highly category-related, and they are unable to track unseen categories.
ColTrack~\cite{liu2023collaborative} leverages multiple historical queries of the same ID to become frame-rate-insensitive.
However, multiple queries add computational burden, and ColTrack also struggles to perform tracking for unknown categories.

\subsection{Open-Vocabulary MOT}
With the advancement of the MOT community, various MOT datasets~\cite{milan2016mot16,dendorfer2020mot20,sun2022dancetrack} have abundant scenarios and a variety of motion patterns.
However, these datasets predominantly feature the person category, which is not in accord with the practical and universal scenario.
To address this problem, OV-MOT aims to track unseen categories.
Unfortunately, very few papers have tackled the OV-MOT problem.
Li et al.~\cite{li2023ovtrack} first define the task of OV-MOT and propose OVTrack to handle this problem on the TAO~\cite{dave2020tao} dataset.
Different from classical CV-MOT (i.e., the categories for testing and training are consistent), OV-MOT splits categories of objects into non-overlapped $\mathcal{C}^{\mathrm{base}}$ and $\mathcal{C}^{\mathrm{novel}}$.
Only $\mathcal{C}^{\mathrm{base}}$ is used for training, and $\mathcal{C}^{\mathrm{base}}$ together with $\mathcal{C}^{\mathrm{novel}}$ is then used for testing by providing names of the novel categories.
OVTrack leverages distilled knowledge from CLIP~\cite{radford2021learning} to track unseen categories $\mathcal{C}^{\mathrm{novel}}$.
Moreover, it employs a contrastive learning approach to distinguish different IDs.
By integrating 10 million automatically labeled data, GLEE~\cite{wu2023general} uses a unified framework to address several open-vocabulary tasks like detection, segmentation, grounding, tracking, etc.
However, these methods overlook the true needs of the model during the association process leading to their suboptimal performance.
NetTrack~\cite{zheng2024nettrack} leverages fine-grained features, i.e., point-level visual cues, to track highly dynamic objects.
Nevertheless, its association performance is still less than ideal in some extremely dynamic scenarios like TAO.

\section{Method}
\label{sec:method}

In this section, we introduce AED that aims to \textbf{A}ssociate \textbf{E}verything \textbf{D}etected.
Given arbitrary detection results, it can track objects in both closed and open vocabulary scenarios.
It contains two key components: similarity decoder (sim-decoder) and association-centric learning mechanism.
We first give an overview of AED in Section~\ref{sec:overview}, and then introduce sim-decoder and association-centric learning mechanism in Section~\ref{sec:sim-decoder} and Section~\ref{sec:association-centric} respectively.
In Section~\ref{sec:training}, we provide some training details of AED.

\begin{figure*}[!t]
\centering
\includegraphics[width=\textwidth]{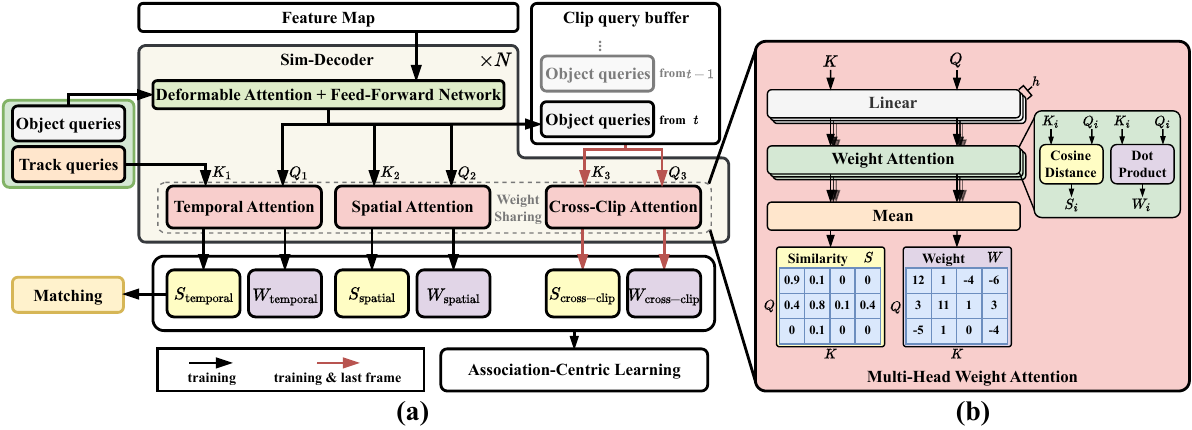}
\caption{Design of (a) similarity decoder (sim-decoder) and (b) multi-head weight attention. Object queries and feature maps of the current frame are fed into a multi-head deformable cross-attention and feed-forward network to extract appearance features of object queries, where the residual connection is omitted for simplicity. During training, we use three multi-head weight attentions (i.e., Temporal/Spatial/Cross-Clip Attention) with shared weights to get temporal similarities $S_{\mathrm{temporal}}$ and weights $W_{\mathrm{temporal}}$, spatial similarities $S_{\mathrm{sptial}}$ and weights $W_{\mathrm{sptial}}$, and cross-clip similarities $S_{\mathrm{cross-clip}}$ and weights $W_{\mathrm{cross-clip}}$ respectively. The temporal similarities $S_{\mathrm{temporal}}$ is then used for matching. These three similarity and weight matrices are then used for the association-centric learning mechanism.}
\label{fig:sim-decoder}
\vspace{-0.7cm}
\end{figure*}

\subsection{Overview}
\label{sec:overview}
We give an overview of AED in Fig.~\ref{fig:overview}.
We build AED upon the tracking-by-detection paradigm for two reasons.
On the one hand, detection results can be arbitrary objects of any category during inference, which is the fundamental precondition of AED.
On the other hand, AED can be applied plug-and-play to any object detectors.

We design AED to decode similarities between detection results and historical trajectories and perform matching.
The input of AED is a video sequence.
For every frame, there is an object detector (e.g., Co-DETR~\cite{codetr}, YOLOX~\cite{ge2021yolox} or RegionCLIP~\cite{regionclip}) outputting bounding boxes and a CNN backbone (i.e., ResNet50~\cite{he2016deep}) followed with an encoder~\cite{zhu2020deformable} extracting image features.
For every frame, we encode the detected boxes into object queries.
Therefore, we regard the detections as object queries and the existing trajectories as track queries of the current frame.
Then the object queries and track queries are fed into a novel sim-decoder (Section~\ref{sec:sim-decoder}) to decode their similarities.
The sim-decoder also aggregates the appearance features of object queries layer by layer.
To minimize the reliance on prior knowledge and adapt to the vast majority of motion patterns, AED leverages only the decoded similarity to accomplish the matching process.

Different from existing tracking-by-query methods~\cite{zeng2022motr, zhang2023motrv2} that track one trajectory by using one query, in AED, the matched track queries will be updated by the corresponding object queries.
Then, the matched object queries and the unmatched track queries are treated as track queries for the next frame.
Note that we do not perform matching at the end of the first frame because no track query exists at that time.

To maximize the consistency between training and online association and also enhance the robustness of AED to generalize to unknown categories and motion patterns, we propose an association-centric learning mechanism (Section~\ref{sec:association-centric}) during training.
Association-centric learning prompts AED to learn complex matching scenarios that are similar to the association phase.
By leveraging the sim-decoder, the association-centric learning mechanism is reflected in three aspects: spatial, temporal, and cross-clip.
Specifically, we apply spatial contrastive learning to make AED better distinguish objects within one single frame, which is the precondition for cross-frame associations.
Then we take a sampled video clip with $n$ frames as the input and use temporal contrastive learning to help AED match objects with historical trajectories.
Moreover, we use cross-clip contrastive learning to improve the long-term ID consistency of AED on several frames.

\subsection{Similarity Decoder}
\label{sec:sim-decoder}

We formulate the learning process of AED as a similarity and weight response decoding problem.
As illustrated in Fig.~\ref{fig:sim-decoder}(a), we propose a similarity decoder (sim-decoder) with $N$ layers to decode spatial, temporal, and cross-clip similarities $S$ and weight responses $W$.

The input of the sim-decoder consists of three parts: object queries, track queries, and features map of the current frame.
Based on the confidence scores output by the detector, we first encode $M_{t}$ bounding boxes from frame $t$ into object queries of size $M_{t}\times d_{\text{model}}$ by using sine-cosine positional encoding~\cite{vaswani2017attention}, where $d_{\text{model}}$ denotes the feature dimension of each query.
Since the object queries only contain confidence information at this time, we employ a multi-head deformable cross-attention~\cite{zhu2020deformable} and a feed-forward network to aggregate their appearance features according to the feature map.
To help sim-decoder better locate the feature, we employ box refinement using object queries following~\cite{zhang2023motrv2} (omitted in Fig.~\ref{fig:sim-decoder}(a) for simplicity and will be discussed in Section~\ref{sec:training}). We further utilize object queries in three attentional dimensions during training.

\subsubsection{Temporal Attention}
Temporal attention takes track queries and object queries as the input keys ($K_1$) and queries ($Q_1$).
The track queries consist of the matched object queries and the unmatched track queries from the previous frame.
This attention calculates pairwise similarities $S_{\mathrm{temporal}}$ and weight responses $W_{\mathrm{temporal}}$ between object and track queries.
$S_{\mathrm{temporal}}$ and $W_{\mathrm{temporal}}$ are then employed in association-centric learning (Section~\ref{sec:association-centric}) to guarantee robust matching between detections and trajectories.

\subsubsection{Spatial Attention}
The spatial attention then operates on the object queries within a frame to compute pairwise $S_{\mathrm{spatial}}$ and $W_{\mathrm{spatial}}$ between these queries.
These matrices physically represent the similarity measurements and weights among detections within individual frames.
$S_{\mathrm{spatial}}$ and $W_{\mathrm{spatial}}$ then help to keep the spatial discriminability of AED in association-centric learning.

\subsubsection{Cross-Clip Attention}
During training, all output object queries from the sampled video clip are cached in a clip query buffer.
Upon reaching the terminal frame of the clip, these buffered queries are simultaneously assigned as keys ($K_3$) and queries ($Q_3$) for cross-clip attention computation, generating $S_{\mathrm{cross-clip}}$ and $W_{\mathrm{cross-clip}}$.
These matrices denote the similarities and weight responses among all the objects that appear in this clip.
Through association-centric learning, $S_{\mathrm{cross-clip}}$ and $W_{\mathrm{cross-clip}}$ are further optimized to improve the long-term trajectory modeling capability of AED.

All three attention mechanisms mentioned above are based on our proposed multi-head weight attention, as illustrated in Fig.~\ref{fig:sim-decoder}(b).
Inspired by the ”weight” concept in standard multi-head attention~\cite{vaswani2017attention}, we reformulate it as multi-head weight attention by eliminating the ”value” component, specializing in deriving similarities $S$ and weight responses $W$ between input keys $K$ and queries $Q$.
The more similar a given pair of the key and query (e.g., from the same ID), the higher the corresponding similarity and response will be.
To be specific, we linearly project $K$ and $Q$ for $h$ times respectively, where $h$ is the number of heads.
For each $\text{head}_i$, we compute the similarities $S_i$ and the weight responses $W_i$ between the projected queries and keys using weight attention.
$h$ weight attentions are employed for the similarity learning from different representation subspaces.
For $\text{head}_i$, its $S_i$ and $W_i$ can be formulated as:
\begin{align}
    S_i&=\mathrm{norm}(\mathrm{linear}_{i}(Q))\cdot \mathrm{norm}(\mathrm{linear}_{i}(K))^{T}, \\
    W_i&=\mathrm{linear}_{i}(Q)\cdot \mathrm{linear}_{i}(K)^{T}.
\end{align}
The elements in $S_i$ and $W_i$ are the cosine similarities and dot products between linearly mapped keys and queries respectively.
Different from~\cite{vaswani2017attention}, for $S_i$, we employ cosine similarity instead of dot product with softmax in our weight attention because the softmax function can only identify the relative maximum, not the absolute maximum, which is catastrophic for data association in MOT.
For example, newly emerged objects should have low similarity with all historical trajectories.
If softmax is used, the similarities will be comparatively high.

The results of $h$ weight attentions are then averaged element-wise to output the final $S$ and $W$ for multi-head weight attention (i.e., temporal/spatial/cross-clip attention in Fig.~\ref{fig:sim-decoder}(a)):
\begin{align}
    S&=\mathrm{max}(0, \mathrm{mean}(S_1,S_2,...,S_h)), \\
    W&=\mathrm{mean}(W_1,W_2,...,W_h).
\end{align}
Note that for each element of $S$, we constrain its value within the range of [0,1].

During inference, only $S_{\mathrm{temporal}}$ is required to perform matching, and we take the entire video sequence as the input.
Instead of leveraging some prior knowledge like Kalman filter~\cite{kalman1960new}, we directly use the $S_{\mathrm{temporal}}$ from the final layer of sim-decoder as the input for the Hungarian algorithm~\cite{munkres1957algorithms} to perform matching.
Again, we simply update the track query with the corresponding object query if they match successfully.
Inspired by~\cite{zhang2022bytetrack}, we employ a very simple two-stage matching strategy.
Specifically, we first associate the high-scoring boxes and then proceed to associate the low-scoring boxes.
AED inits every untracked bounding box as a new trajectory without any complex initialization procedures like~\cite{wojke2017simple, zhang2022bytetrack, li2023ovtrack}.

\subsection{Association-Centric Learning Mechanism}
\label{sec:association-centric}

Based on the three sets of $S$ and $W$ that correspond to spatial, temporal, and cross-clip, we propose an association-centric learning mechanism to tackle the aforementioned association-learning ignorance problem and keep the extracted features robust enough to generalize to the unknown categories.
The association-centric learning model the training process as a contrastive learning.

\begin{figure*}[!t]
  \centering
  \includegraphics[height=9cm]{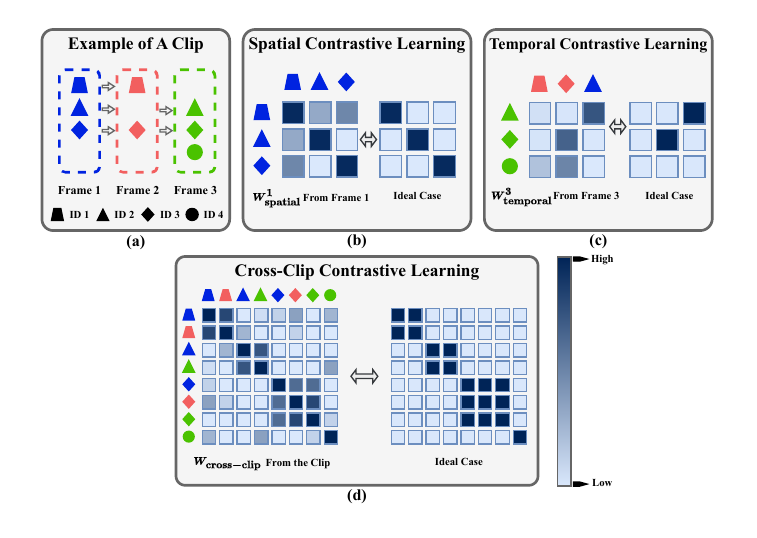}
  \caption{Visualization of association-centric learning mechanism. It includes three aspects: spatial contrastive learning, temporal contrastive learning, and cross-clip contrastive learning.
  }
  \label{fig:association-centric}
  \vspace{-0.5cm}
\end{figure*}

To better understand our approach, we visualize the association-centric learning in Fig.~\ref{fig:association-centric}.
Fig.~\ref{fig:association-centric}(a) provides an example of a sampled clip containing three frames during training.
Assuming that there are four IDs in this clip, ID 1 and ID 2 are absent in frame 3 and frame 2 respectively.
ID 4 only appears in frame 3.
During training, their corresponding object queries are encoded from real detections.

For each $W$ (i.e., $W_{\mathrm{spatial/temporal/cross-clip}}$) output by sim-decoder, we apply the embedding loss proposed in~\cite{pang2021quasi} to enhance the weight responses among keys and queries that share the same ID and diminish the responses from different IDs.
We formulate the loss as follows:
\begin{equation}
\mathcal{L}_{\mathrm{embed}}(W)=\sum_{r}\log[1+\sum_{w_{r}^{+}}\sum_{w_{r}^{-}}\exp(w_{r}^{-}-w_{r}^{+})],
\end{equation}
where $r$ is the index of rows in $W$, and each row of $W$ represents the responses of all keys to the same query in multi-head weight attention.
$w^{+}_{r},w^{-}_{r}\in W_{r}$ and $W_{r}$ denotes the $r$-th row of $W$.
$w^{+}_{r}$ is the response between key and query from the same ID, and $w^{-}_{r}$ is the response between different IDs.
We further adopt focal loss~\cite{lin2017focal} as the auxiliary loss based on the corresponding $S$ (i.e., $S_{\mathrm{spatial/temporal/cross-clip}}$):
\begin{equation}
\mathcal{L}_{\mathrm{aux}}(S)=-\sum_{s^{+}}(1-s^{+})^\gamma\log(s^{+})-\sum_{s^{-}}(s^{-})^\gamma\log(1-s^{-}),
\end{equation}
where $s^{+},s^{-}\in S$ and $s^{+}$ is the similarity between the same ID, $s^{-}$ is the similarity between different IDs.

We argue that distinguishing different IDs within a single frame is fundamental for a tracker before cross-frame data association.
Additionally, we all know that objects within a single frame are unique.
As shown in Fig.~\ref{fig:association-centric}(b), a target should only have a high response with itself within a single frame.
Thus, we leverage $W_{\mathrm{spatial}}$ and $S_{\mathrm{spatial}}$ to enhance the spatial discriminative ability of AED.
Specifically, we use $\mathcal{L}_{\mathrm{embed}}$ and $\mathcal{L}_{\mathrm{aux}}$ to accomplish spatial contrastive learning:
\begin{equation}
\mathcal{L}_{\mathrm{spatial}}=\sum_{t}(\mathcal{L}_{\mathrm{embed}}(W_{\mathrm{spatial}}^{t})+\mathcal{L}_{\mathrm{aux}}(S_{\mathrm{spatial}}^{t})),
\end{equation}
where $t$ is the frame index from the sampled clip.

Temporal contrastive learning aims to connect object queries with track queries that share the same ID.
During training, we perform matching to generate real track queries, rather than relying on information from two joint frames. This can be seen in Fig.~\ref{fig:association-centric}(c), where ID2 from frame 1 is also included.
Thus, by using $W_{\mathrm{temporal}}$ and $S_{\mathrm{temporal}}$, temporal contrastive learning can be formulated as:
\begin{equation}
\mathcal{L}_{\mathrm{temporal}}=\sum_{t}(\mathcal{L}_{\mathrm{embed}}(W_{\mathrm{temporal}}^{t})+\mathcal{L}_{\mathrm{aux}}(S_{\mathrm{temporal}}^{t})).
\end{equation}

Since the ultimate object of MOT is to keep the ID as long as possible, we further use cross-clip contrastive learning to intensify the long-term discriminability of AED.
As shown in Fig.~\ref{fig:association-centric}(d), we use all the objects in this clip to learn the feature consistency of targets across multiple frames.
Similarly, for the entire clip, we define the cross-clip contrastive learning as:
\begin{equation}
\mathcal{L}_{\mathrm{cross-clip}}=\mathcal{L}_{\mathrm{embed}}(W_{\mathrm{cross-clip}})+\mathcal{L}_{\mathrm{aux}}(S_{\mathrm{cross-clip}}).
\end{equation}
Instead of introducing an additional memory module \cite{cai2022memot,gao2023memotr}, our method integrates long-term modeling directly into the training process and achieves significant computational savings.

To sum up, association-centric learning aligns the training process with the association process, making AED more robust to different tracking scenarios.

\subsection{Training Details}
\label{sec:training}
As mentioned in Section~\ref{sec:sim-decoder}, we refine the boxes during training.
In detail, we use the object queries output by the feed-forward network in the sim-decoder and apply an additional multi-layer perceptron (MLP) to predict the compensation relative to their ground truth (GT) boxes, following~\cite{zhang2023motrv2}.
The box refinement loss includes L1 loss and GIoU loss~\cite{rezatofighi2019generalized} denoted as $\mathcal{L}_{\mathrm{l_1}}$ and $\mathcal{L}_{\mathrm{giou}}$ respectively.

As a result, our total training loss is formulated as follows:
\begin{equation}
    \begin{aligned}
        \mathcal{L}_{\mathrm{total}}&=\lambda_{\mathrm{spatial}}\mathcal{L}_{\mathrm{spatial}}+\lambda_{\mathrm{temporal}}\mathcal{L}_{\mathrm{temporal}} \\
                                     &+\lambda_{\mathrm{cross-clip}}\mathcal{L}_{\mathrm{cross-clip}}+\lambda_{\mathrm{l1}}\mathcal{L}_{\mathrm{l1}} \\
                                     &+\lambda_{\mathrm{giou}}\mathcal{L}_{\mathrm{giou}},
    \end{aligned}
\end{equation}
where $\lambda_{\mathrm{spatial}}$, $\lambda_{\mathrm{temporal}}$, $\lambda_{\mathrm{cross-clip}}$, $\lambda_{\mathrm{l1}}$, and $\lambda_{\mathrm{giou}}$ are the hyperparameters of their corresponding loss functions, and we set them to 0.1, 2, 1, 0.5, and 0.3 respectively.
To speed up convergence and enhance the precision of the model, we also employ auxiliary losses~\cite{al2019character} at each layer of our sim-decoder.

We use real detected boxes to generate object queries during training including the boxes that are not matched with any GT, e.g., false positives.
The matched boxes (i.e., IoU with GT $>0.5$) will contribute to all of the loss functions.
The unmatched boxes will contribute to $\mathcal{L}_{\mathrm{spatial}}$ and will act as negative samples in $\mathcal{L}_{\mathrm{temporal}}$.
If any object query is matched with a GT initially but its IoU with GT is smaller than a threshold $\tau_{\mathrm{iou}}$ after box refinement, we simply filter it out to avoid ambiguous learning.

For OV-MOT tasks, we split categories into $\mathcal{C}^{\mathrm{base}}$ and $\mathcal{C}^{\mathrm{novel}}$ following~\cite{li2023ovtrack}.
Specifically, $\mathcal{C}^{\mathrm{novel}}$ are the rare classes defined in \cite{gupta2019lvis} and $\mathcal{C}^{\mathrm{base}}$ are the rest of the classes in TAO.
We only use $\mathcal{C}^{\mathrm{base}}$ to train AED, and we use $\mathcal{C}^{\mathrm{base}}$ and $\mathcal{C}^{\mathrm{novel}}$ to evaluate our models.
Introducing new categories during testing that are not seen in the training phase can evaluate the class generalization ability of the model.
For CV-MOT tasks, we utilize the entirety of the annotated data for training.

\section{Experiments}

\begin{table*}[!t]
    \caption{Comparison with state-of-the-art OV-MOT trackers on TAO dataset.
             We categorize the experimental results based on the differences in detectors.
             "Integrated" indicates this method uses its model to localize and extract features of objects simultaneously.
              The best results are shown in \textbf{bold} and the second best results are shown with \underl{underlines}.}
    \label{tab:ovmot_comparison}
    \centering
        \begin{tabularx}{\textwidth}
        {l|l|l|*{4}{>{\centering\arraybackslash}X}|*{4}{>{\centering\arraybackslash}X}}
        \toprule
        \mc{1}{c|}{\mr{2}{*}{Detector}}         & \mc{2}{c|}{Method}                                & \mc{4}{c|}{Base}                                               & \mc{4}{c}{Novel}                                               \\
        \cmidrule{2-11}
                                                & Validation set                        &Publication&TETA$\uparrow$ &LocA$\uparrow$ &AssocA$\uparrow$&ClsA$\uparrow$ &TETA$\uparrow$ & LocA$\uparrow$&AssocA$\uparrow$&ClsA$\uparrow$ \\
        \midrule
        \mr{3}{*}{Integrated}                   & QDTrack~\cite{pang2021quasi}           & CVPR2021 & 27.1          & 45.6          & 24.7           & 11.0          & 22.5          & 42.7          & 24.4           & 0.4           \\
                                                & TETer~\cite{li2022tracking}            & ECCV2022 & 30.3          & 47.4          & 31.6           & 12.1          & 25.7          & 45.9          & 31.1           & 0.2           \\
                                                & OVTrack~\cite{li2023ovtrack}           & CVPR2023 & 35.5          & 49.3          & 36.9           & 20.2          & 27.8          & 48.8          & 33.6           & 1.5           \\
        \midrule
        \mr{1}{*}{GroundingDINO}~\cite{liu2023grounding}
                                                & NetTrack~\cite{zheng2024nettrack}      & CVPR2024 & 33.0          & 45.7          & 28.6           & 24.8          & 32.6          & 51.3          & 33.0           & 13.3          \\
        \midrule
        \mr{6}{*}{RegionCLIP~\cite{regionclip}} & DeepSORT~\cite{wojke2017simple}        & ICIP2017 & 28.4          & 52.5          & 15.6           & 17.0          & 24.5          & 49.2          & 15.3           & 9.0           \\ %
                                                & ByteTrack~\cite{zhang2022bytetrack}    & ECCV2022 & 29.4          & 52.3          & 19.8           & 16.0          & 26.5          & 50.8          & 20.9           & 8.0           \\ %
                                                & Tracktor++~\cite{bergmann2019tracking} & ICCV2019 & 29.6          & 52.4          & 19.6           & 16.9          & 25.7          & 50.1          & 18.9           & 8.1           \\
                                                & OVTrack~\cite{li2023ovtrack}           & CVPR2023 & 36.3          & 53.9          & 36.3           & 18.7          & 32.0          & 51.4          & 33.2           & 11.4          \\
                                                & AED                                    & Ours     & 37.3          & 56.4          & 38.4           & 17.0          & 34.4          & 55.9          & 38.5           & 8.7           \\ %
        \midrule
        \mr{4}{*}{Co-DETR~\cite{codetr}}        & DeepSORT~\cite{wojke2017simple}        & ICIP2017 & 45.8          & \underl{70.5} & 28.1           & 38.9          & 42.2          & 72.7          & 29.0           & 24.7          \\ %
                                                & ByteTrack~\cite{zhang2022bytetrack}    & ECCV2022 & 41.0          & 62.3          & 15.9           & \underl{44.8} & 37.2          & 65.9          & 17.0           & \textbf{28.7} \\ %
                                                & OVTrack~\cite{li2023ovtrack}           & CVPR2023 & \underl{54.5} & 70.3          & \underl{47.5}  & \textbf{45.6} & \underl{49.1} & \underl{73.7} & \underl{46.3}  & 27.2          \\ %
                                                & AED                                    & Ours     & \textbf{55.7} & \textbf{71.5} & \textbf{52.2}  & 43.5          & \textbf{51.4} & \textbf{74.4} & \textbf{51.8}  & \underl{28.1} \\ %
        \bottomrule
        \toprule
        \mc{1}{c|}{Detector}                    & Test set                              &Publication&TETA$\uparrow$ &LocA$\uparrow$ &AssocA$\uparrow$&ClsA$\uparrow$ &TETA$\uparrow$ & LocA$\uparrow$&AssocA$\uparrow$&ClsA$\uparrow$ \\
        \midrule
        \mr{3}{*}{Integrated}                   & QDTrack~\cite{pang2021quasi}           & CVPR2021 & 25.8          & 43.2          & 23.5           & 10.6          & 20.2          & 39.7          & 20.9           & 0.2           \\
                                                & TETer~\cite{li2022tracking}            & ECCV2022 & 29.2          & 44.0          & 30.4           & 10.7          & 21.7          & 39.1          & 25.9           & 0.0           \\
                                                & OVTrack~\cite{li2023ovtrack}           & CVPR2023 & 32.6          & 45.6          & 35.4           & 16.9          & 24.1          & 41.8          & 28.7           & 1.8           \\
        \midrule
        \mr{5}{*}{RegionCLIP~\cite{regionclip}} & DeepSORT~\cite{wojke2017simple}        & ICIP2017 & 27.0          & 49.8          & 15.1           & 16.1          & 18.7          & 41.8          & 9.1            & 5.2           \\ %
                                                & Tracktor++~\cite{bergmann2019tracking} & ICCV2019 & 28.0          & 49.4          & 18.8           & 15.7          & 20.0          & 42.4          & 12.0           & 5.7           \\
                                                & ByteTrack~\cite{zhang2022bytetrack}    & ECCV2022 & 28.7          & 51.5          & 19.9           & 14.5          & 20.4          & 43.0          & 13.5           & 4.9           \\ %
                                                & OVTrack~\cite{li2023ovtrack}           & CVPR2023 & 34.8          & 51.1          & 36.1           & 17.3          & 25.7          & 44.8          & 26.2           & 6.1           \\
                                                & AED                                    & Ours     & 37.2          & 55.7          & 40.4           & 15.6          & 27.8          & 48.5          & 29.1           & 5.9           \\ %
        \midrule
        \mr{4}{*}{Co-DETR~\cite{codetr}}        & DeepSORT~\cite{wojke2017simple}        & ICIP2017 & 45.6          & \underl{69.7} & 30.3           & 36.9          & 41.1          & \underl{70.3} & 27.8           & \underl{25.1} \\ %
                                                & ByteTrack~\cite{zhang2022bytetrack}    & ECCV2022 & 38.8          & 60.6          & 14.8           & \underl{41.1} & 31.4          & 55.4          & 15.1           & 23.7          \\ %
                                                & OVTrack~\cite{li2023ovtrack}           & CVPR2023 & \underl{53.6} & 69.5          & \underl{49.3}  & \textbf{42.1} & \underl{45.9} & 66.0          & \underl{45.3}  & \textbf{26.3} \\ %
                                                & AED                                    & Ours     & \textbf{54.8} & \textbf{70.6} & \textbf{54.1}  & 39.9          & \textbf{48.9} & \textbf{70.5} & \textbf{51.8}  & 24.6          \\ %
        \bottomrule
        \end{tabularx}
        \vspace{-0.2cm}
\end{table*}

\subsection{Datasets And Evaluation Metrics}
\noindent \textbf{Datasets.}
We use TAO~\cite{dave2020tao}, DanceTrack~\cite{sun2022dancetrack}, and SportsMOT~\cite{cui2023sportsmot} datasets to evaluate both OV-MOT and CV-MOT performance of AED.
TAO is a large-scale dataset that contains 482 categories for MOT tasks.
It is annotated under the federated protocol~\cite{gupta2019lvis}, and there are up to 10 labeled objects in every video of TAO.
Additionally, TAO is annotated at a quite low frame rate, i.e., 1 FPS.
SportsMOT aims to track players on the court, including basketball, volleyball, and football.
Except for similar appearances, SportsMOT also has fast and variable-speed motion, which poses a higher challenge to the performance of the tracker.
DanceTrack also has one category: person.
It is a challenging dataset with diverse motion and severe occlusion, and people in DanceTrack also have very similar appearances.

\noindent \textbf{Evaluation Metrics.}
AED can apply to both OV-MOT and CV-MOT tasks.
Therefore, we use Tracking-Every-Thing Accuracy (TETA)~\cite{li2022tracking} when comparing AED with other OV-MOT trackers.
For CV-MOT tasks, i.e., DanceTrack and SportsMOT, we take HOTA~\cite{luiten2021hota}, MOTA~\cite{bernardin2008evaluating}, and IDF1~\cite{ristani2016performance} as the main evaluation metrics.

\subsection{Implementation Details}
By default, we use ResNet50~\cite{he2016deep} as the CNN backbone of AED, and we have 6 layers of encoder and sim-decoder.
We initialize AED with the Deformable DETR~\cite{zhu2020deformable} parameters pretrained on the COCO dataset~\cite{lin2014coco}.
We use AdamW optimizer~\cite{loshchilov2017decoupled} to train AED for 5 epochs on a single RTX 4090 GPU.
The initial learning rate is $1\times 10^{-4}$, and it decreases by a factor of 10 every 2 epochs.
Each batch consists of a clip sampled from a video sequence.
We set the number of sampled frames $n$ in a clip to be 5.
The $d_{\text{model}}$ is set to 256.
For the multi-head weight attention, $h$ is set to 2.

We assign each detection box to a GT box based on an IoU threshold ($>0.5$) during the clip sampling process.
Since the TAO dataset is annotated under the federated protocol~\cite{gupta2019lvis}, i.e., not all instances of the same category are annotated in the images, the treatment of these unmatched boxes follows Section~\ref{sec:training}.

During inference, we do not perform box refinement for OV-MOT tasks to avoid inaccurate refinement for the unseen $\mathcal{C}^{\mathrm{novel}}$.
The output categories are directly inherited from the detector.

\begin{table*}[!t]
    \caption{Comparison with state-of-the-art CV-MOT trackers on SportsMOT dataset.
             Methods in the \textcolor{gray}{gray} blocks use the same detector (YOLOX).
             AED$^{*}$ denotes AED is trained on the $\mathcal{C}^{\mathrm{base}}$ of TAO dataset following Section~\ref{sec:exp_ovmot} while the training data of the detector follows the "Train Setup".
             The best results are shown in \textbf{bold} and the second best results are shown with \underl{underlines}.}
    \label{tab:sportsmot_comparison}
    \centering
        \begin{tabularx}{\textwidth}
        {l|l|c|*{7}{>{\centering\arraybackslash}X}}
        \toprule
        \multicolumn{1}{c|}{Methods}                          & Publication & Training Setup  & HOTA$\uparrow$& IDF1$\uparrow$& AssA$\uparrow$& MOTA$\uparrow$& DetA$\uparrow$& LocA$\uparrow$& IDs$\downarrow$\\
        \midrule
        FairMOT~\cite{zhang2021fairmot}                       & IJCV2021    & Train           & 49.3          & 53.5          & 34.7          & 86.4          & 70.2          & 83.9          & 9928           \\
        QDTrack~\cite{pang2021quasi}                          & CVPR2021    & Train           & 60.4          & 62.3          & 47.2          & 90.1          & 77.5          & 88.0          & 6377           \\
        CenterTrack~\cite{zhou2020centertrack}                & ECCV2020    & Train           & 62.7          & 60.0          & 48.0          & 90.8          & 82.1          & 90.8          & 10481          \\
        TransTrack~\cite{sun2020transtrack}                   & arXiv2020   & Train           & 68.9          & 71.5          & 57.5          & 92.6          & 82.7          & 91.0          & 4992           \\
        \rowcolor{gray!20}ByteTrack~\cite{zhang2022bytetrack} & ECCV2022    & Train           & 62.8          & 69.8          & 51.2          & 94.1          & 77.1          & 85.6          & 3267           \\
        \rowcolor{gray!20}BoT-SORT~\cite{aharon2022bot}       & arXiv2022   & Train           & 68.7          & 70.0          & 55.9          & 94.5          & 84.4          & 90.5          & 6729           \\
        \rowcolor{gray!20}OC-SORT~\cite{cao2023observation}   & CVPR2023    & Train           & 71.9          & 72.2          & 59.8          & 94.5          & 86.4          & \underl{92.4} & 3093           \\
        \rowcolor{gray!20}DiffMOT~\cite{lv2024diffmot}        & CVPR2024    & Train           & 72.1          & 72.8          & 60.5          & 94.5          & 86.0          & -             & -              \\
        \rowcolor{gray!20}Deep-EIoU~\cite{huang2024iterative} & WACV2024    & Train           & \underl{74.1} & 75.0          & \underl{63.1} & \textbf{95.1} & \textbf{87.2} & \textbf{92.5} & 3066           \\
        \rowcolor{gray!20}AED$^{*}$                           & Ours        & Train           & 72.8          & \underl{76.8} & 61.4          & \underl{95.0} & 86.3          & 92.0          & \underl{2607}  \\
        \rowcolor{gray!20}AED                                 & Ours        & Train           & \textbf{77.0} & \textbf{80.0} & \textbf{68.1} & \textbf{95.1} & \underl{87.1} & \textbf{92.5} & \textbf{2240}  \\
        \midrule
        \rowcolor{gray!20}ByteTrack~\cite{zhang2022bytetrack} & ECCV2022    & Train+Val       & 64.1          & 71.4          & 52.3          & 95.9          & 78.5          & 85.7          & 3089           \\
        \rowcolor{gray!20}MixSort-Byte~\cite{cui2023sportsmot}& ICCV2023    & Train+Val       & 65.7          & 74.1          & 54.8          & 96.2          & 78.8          & 85.7          & 2472           \\
        \rowcolor{gray!20}OC-SORT~\cite{cao2023observation}   & CVPR2023    & Train+Val       & 73.7          & 74.0          & 61.5          & 96.5          & 88.5          & \underl{92.7} & 2728           \\
        \rowcolor{gray!20}MixSort-OC~\cite{cui2023sportsmot}  & ICCV2023    & Train+Val       & 74.1          & 74.4          & 62.0          & 96.5          & 88.5          & \underl{92.7} & 2781           \\
        \rowcolor{gray!20}DiffMOT~\cite{lv2024diffmot}        & CVPR2024    & Train+Val       & 76.2          & 76.1          & 65.1          & \textbf{97.1} & \underl{89.3} & -             & -              \\
        \rowcolor{gray!20}Deep-EIoU~\cite{huang2024iterative} & WACV2024    & Train+Val       & \underl{77.2} & \underl{79.8} & \underl{67.7} & 96.3          & 88.2          & 92.4          & 2659           \\
        \rowcolor{gray!20}AED$^{*}$                           & Ours        & Train+Val       & 74.4          & 78.2          & 62.8          & \underl{97.0} & 88.2          & 92.2          & \underl{2250}  \\
        \rowcolor{gray!20}AED                                 & Ours        & Train+Val       & \textbf{79.1} & \textbf{81.8} & \textbf{70.1} & \textbf{97.1} & \textbf{89.4} & \textbf{92.8} & \textbf{1855}  \\
        \bottomrule
        \end{tabularx}
        \vspace{-0.2cm}
\end{table*}

\begin{table*}[!t]
    \caption{Comparison with state-of-the-art CV-MOT trackers on DanceTrack dataset.
             Methods in the \textcolor{gray}{gray} blocks use the same detector (YOLOX).
             AED$^{*}$ denotes AED is trained on the $\mathcal{C}^{\mathrm{base}}$ of TAO dataset following Section~\ref{sec:exp_ovmot}.
             The best results are shown in \textbf{bold} and the second best results are shown with \underl{underlines}.}
    \label{tab:dancetrack_comparison}
    \centering
        \begin{tabularx}{\textwidth}
        {l|l|*{5}{>{\centering\arraybackslash}X}}
        \toprule
        \multicolumn{1}{c|}{Methods}                           & Publication & HOTA$\uparrow$ & IDF1$\uparrow$ & AssA$\uparrow$ & MOTA$\uparrow$ & DetA$\uparrow$\\
        \midrule
        {\color{gray} Tracking-by-Detection Methods:}                                   & & & & & \\
        FairMOT~\cite{zhang2021fairmot}                        & IJCV2021    & 39.7           & 40.8          & 23.8           & 82.2           & 66.7          \\
        CenterTrack~\cite{zhou2020centertrack}                 & ECCV2020    & 41.8           & 35.7          & 22.6           & 86.8           & 78.1          \\
        QDTrack~\cite{pang2021quasi}                           & CVPR2021    & 45.7           & 44.8          & 29.2           & 83.0           & 72.1          \\
        GTR~\cite{zhou2022global}                              & CVPR2022    & 48.0           & 50.3          & 31.9           & 84.7           & 72.5          \\
        FineTrack~\cite{ren2023focus}                          & CVPR2023    & 52.7           & 59.8          & 38.5           & 89.9           & 72.4          \\
        \rowcolor{gray!20} DeepSORT~\cite{wojke2017simple}     & ICIP2017    & 45.6           & 47.9          & 29.7           & 87.8           & 71.0          \\
        \rowcolor{gray!20} ByteTrack~\cite{zhang2022bytetrack} & ECCV2022    & 47.3           & 52.5          & 31.4           & 89.5           & 71.6          \\
        \rowcolor{gray!20} SORT~\cite{bewley2016simple}        & ICIP2016    & 47.9           & 50.8          & 31.2           & 91.8           & 72.0          \\
        \rowcolor{gray!20} OC-SORT~\cite{cao2023observation}   & CVPR2023    & 54.6           & 54.6          & 40.2           & 89.6           & 80.4          \\
        \rowcolor{gray!20} DiffMOT~\cite{lv2024diffmot}   & CVPR2024    & 62.3           & 63.0          & \underl{47.2}  & \textbf{92.8}  & \textbf{82.5} \\
        \rowcolor{gray!20} Hybrid-SORT~\cite{yang2024hybrid}   & AAAI2024    & \underl{65.7}  & \underl{67.4} & -              & 91.8           & -             \\
        \rowcolor{gray!20} AED$^{*}$                           & Ours        & 55.2           & 57.0          & 37.8           & 91.0           & 80.8          \\ %
        \rowcolor{gray!20} AED                                 & Ours        & \textbf{66.6}  & \textbf{69.7} & \textbf{54.3}  & \underl{92.2}  & \underl{82.0} \\ %
        \midrule
        {\color{gray} Tracking-by-Query Methods:}                                  & & & & &      \\
        MOTR~\cite{zeng2022motr}                               & ECCV2022    & 54.2           & 51.5          & 40.2           & 79.7           & 73.5         \\
        SUSHI~\cite{cetintas2023unifying}                      & CVPR2023    & 63.3           & 63.4          & 50.1           & 88.7           & 80.1         \\
        MOTRv2~\cite{zhang2023motrv2}                          & CVPR2022    & 69.9           & 71.7          & 59.0           & 91.9           & 83.0         \\
        ColTrack~\cite{liu2023collaborative}                   & ICCV2023    & 72.6           & 74.0          & 62.3           & 92.1           & -            \\
        \bottomrule
        \end{tabularx}
        \vspace{-0.2cm}
\end{table*}

\begin{table}[t]
    \caption{Comparison with state-of-the-art trackers under the closed-vocabulary MOT setting on TAO dataset.
    The best results are shown in \textbf{bold} and the second best results are shown with \underl{underlines}.}
    \label{tab:ovmot_comparison_overall}
    \centering
        \begin{tabularx}{9cm}
        {l|l|*{4}{>{\centering\arraybackslash}X}}
        \toprule
        \mc{1}{c|}{Methods}                   & Publication & TETA$\uparrow$& LocA$\uparrow$& AssocA$\uparrow$& ClsA$\uparrow$\\
        \midrule
        Tracktor~\cite{bergmann2019tracking}  & ICCV2019    & 24.2          & 47.4          & 13.0            & 12.1          \\
        DeepSORT~\cite{wojke2017simple}       & ICIP2017    & 26.0          & 48.4          & 17.5            & 12.1          \\
        Tracktor++~\cite{bergmann2019tracking}         & ECCV2022    & 28.0          & 49.0          & 22.8            & 12.1          \\
        QDTrack~\cite{pang2021quasi}          & CVPR2021    & 30.0          & 50.5          & 27.4            & 12.1          \\
        UNINEXT (R50)~\cite{yan2023universal} & CVPR2023    & 31.9          & 43.3          & 35.5            & 17.1          \\
        TETer~\cite{li2022tracking}           & ECCV2022    & 33.3          & 51.6          & 35.0            & 13.2          \\
        OVTrack~\cite{li2023ovtrack}          & CVPR2023    & 34.7          & 49.3          & 36.7            & 18.1          \\
        GLEE-Pro~\cite{wu2023general}         & CVPR2024    & \underl{47.2} & \underl{66.2} & \underl{46.2}   & \underl{29.1} \\
        \midrule
        AED(RegionCLIP)                       & Ours        & 37.0          & 56.7          & 38.1            & 16.2          \\ %
        AED(Co-DETR)                          & Ours        & \textbf{55.3} & \textbf{71.8} & \textbf{52.4}   & \textbf{41.7} \\ %
        \bottomrule
        \end{tabularx}
        \vspace{-0.2cm}
\end{table}

\subsection{Comparison With OV-MOT Trackers}
\label{sec:exp_ovmot}

In Tab.~\ref{tab:ovmot_comparison}, we give the evaluation result on both validation and test sets of TAO following~\cite{li2023ovtrack}.
AED is trained on $\mathcal{C}^{\mathrm{base}}$ categories on TAO and the object queries are generated from Co-DETR~\cite{codetr}.
During inference, we use two off-the-shelf detectors, RegionCLIP (RN50 backbone version)~\cite{regionclip} and Co-DETR~\cite{codetr}, to provide bounding boxes for AED.
To ensure fair comparison, we use the same detector to evaluate different algorithms.
Due to the categories of the TAO dataset are consistent with those of LVIS~\cite{gupta2019lvis}, RegionCLIP is trained on LVIS with $\mathcal{C}^{\mathrm{base}}$, and Co-DETR is trained on the full set of LVIS.
We simply use a class-agnostic NMS (Non-Maximum Suppression) to remove duplicate detections.

When using RegionCLIP as the detector, AED achieves outstanding performance under the TETA metric compared with several high-performance methods~\cite{pang2021quasi,li2022tracking,li2023ovtrack,zheng2024nettrack,wojke2017simple,zhang2022bytetrack,bergmann2019tracking}.
The previous state-of-the-art method, OVTrack~\cite{li2023ovtrack}, uses a dual contrastive training strategy and achieves excellent performance.
AED outperforms OVTrack by 2.8\% (36.3\%-37.3\%) on TETA of base categories and 7.5\% (32.0\%-34.4\%) on TETA of the unseen novel categories in the validation set of TAO.
When it comes to the test set, AED also has an increment of 6.9\% (34.8\%-37.2\%) on TETA of base categories and 8.2\% (25.7\%-27.8\%) on TETA of the novel categories in the validation set of TAO.
Since AED is responsible for associating every detected bounding box, the AssocA metric directly reflects the performance of AED.
AED outperforms OVTrack by 5.8\% (36.3\%-38.4\%) and 16.0\% (33.2\%-38.5\%) on AssocA of base and novel categories of validation set respectively.
For the test set, the increment for AssocA on base and novel categories are 11.9\% (36.1\%-40.4\%) and 11.1\% (26.2\%-29.1\%) respectively.

Using the Co-DETR detector, AED achieves superior performance among all methods.
Different from RegionCLIP, Co-DETR is a closed-vocabulary detector but AED is trained on $\mathcal{C}^{\mathrm{base}}$ only.
AED surpasses OVTrack on TETA metric by 2.2\% (54.5\%-55.7\%) and 4.7\% (49.1\%-51.4\%) on base and novel categories of validation set respectively, and for the test set, the increment is 2.2\% (53.6\%-54.8\%) and 6.5\% (45.9\%-48.9\%) respectively.
AED also outperforms OVTrack in AssocA by 9.9\% (47.5\%-52.2\%) and 11.9\% (46.3\%-51.8\%) on base and novel categories of the validation set respectively, and in the test set, the increment is 9.7\% (49.3\%-54.1\%) and 14.3\% (45.3\%-51.8\%).

\subsection{Comparison With CV-MOT Trackers}

We use DanceTrack~\cite{sun2022dancetrack}, SportsMOT~\cite{cui2023sportsmot}, and TAO~\cite{dave2020tao} to evaluate the CV-MOT performance of our method.

\subsubsection{SportsMOT}
We further validate the CV-MOT performance of AED on the SportsMOT dataset.
As shown in Tab~\ref{tab:sportsmot_comparison}, for AED, we first use the weights trained on the $\mathcal{C}^{\mathrm{base}}$ of TAO following Section~\ref{sec:exp_ovmot} while the detector (YOLOX~\cite{ge2021yolox}) is trained on the training set of SportsMOT.
Under such a setting, AED achieves a competitive HOTA score, i.e., 72.8, outperforming a large number of trackers~\cite{zhang2021fairmot,pang2021quasi,zhou2020centertrack,sun2020transtrack,aharon2022bot,zhang2022bytetrack,cao2023observation,lv2024diffmot}.
For a more equitable comparison, we also train AED, which is pretrained on COCO, on the training set of SportsMOT.
AED achieves state-of-the-art performance on most metrics.
Compared with Deep-EIoU~\cite{huang2024iterative}, AED achieves the increment of 3.9\% (74.1\%-77.0\%) on HOTA and 6.7\% (75.0\%-80.0\%) on IDF1 by using only training set.
Note that, unlike other tracking-by-detection methods leveraging some extra prior knowledge, AED only uses the similarities of features to achieve this performance.

\subsubsection{DanceTrack}
As shown in Tab.~\ref{tab:dancetrack_comparison}, AED achieves the highest score in HOTA, AssA, and IDF1 metrics compared with some existing state-of-the-art tracking-by-detection CV-MOT methods.
Compared with Hybrid-SORT~\cite{yang2024hybrid}, AED achieves the increment of 1.4\% (65.7\%-66.6\%) on HOTA and 3.4\% (67.4\%-69.7\%) on IDF1 score.
The tracking-by-query CV-MOT methods like MOTRv2~\cite{zhang2023motrv2} and ColTrack~\cite{liu2023collaborative} achieve top-tier performance by tracking objects end-to-end.
However, AED still surpasses some of the tracking-by-query methods like MOTR~\cite{zeng2022motr} and SUSHI~\cite{cetintas2023unifying}.
When using the weights trained on the $\mathcal{C}^{\mathrm{base}}$ of TAO following Section~\ref{sec:exp_ovmot}, AED also achieves competitive performance, i.e., 55.2 HOTA score.

\subsubsection{TAO}
We also validate the CV-MOT performance of AED on the TAO dataset.
We train AED using both $\mathcal{C}^{\mathrm{base}}$ and $\mathcal{C}^{\mathrm{novel}}$ categories.
The results are shown in Tab.~\ref{tab:ovmot_comparison_overall}.
Using Co-DETR as the detector, AED achieves superior performance compared with existing first-class methods including GLEE~\cite{wu2023general} and OVTrack~\cite{li2023ovtrack}.
Moreover, AED also achieves competitive performance by using RegionCLIP.

\subsection{Ablation Studies}

\begin{table}[t]
    \caption{association-centric learning ablation on the validation set of TAO dataset.}
    \label{tab:asscociation-centric_ablation}
    \centering
        \begin{tabularx}{9cm}
        {*{3}{>{\centering\arraybackslash}X}|*{4}{>{\centering\arraybackslash}p{0.6cm}}}
        \toprule
        $\mathcal{L}_{\mathrm{spatial}}$ & $\mathcal{L}_{\mathrm{temporal}}$ & $\mathcal{L}_{\mathrm{cross-clip}}$ & TETA         & LocA          & AssocA          & ClsA          \\
        \midrule
        \checkmark                       &                                  &                                     & 39.5          & 70.4          & 6.4             & \textbf{41.7} \\ %
                                         & \checkmark                       &                                     & 55.0          & \textbf{71.8} & 51.5            & 41.6          \\ %
                                         &                                  & \checkmark                          & 53.0          & \textbf{71.8} & 45.4            & \textbf{41.7} \\ %
                                         & \checkmark                       & \checkmark                          & 55.0          & \textbf{71.8} & 51.4            & \textbf{41.7} \\ %
        \checkmark                       & \checkmark                       &                                     & 55.1          & \textbf{71.8} & 51.8            & 41.6          \\ %
        \checkmark                       &                                  & \checkmark                          & 53.6          & 71.7          & 47.6            & \textbf{41.7} \\ %
        \checkmark                       & \checkmark                       & \checkmark                          & \textbf{55.2} & \textbf{71.8} & \textbf{52.1}   & \textbf{41.7} \\ %
        \bottomrule
        \end{tabularx}
        \vspace{-0.2cm}
\end{table}

In this section, we conduct several ablation experiments to analyze the impact of some components of AED.

\subsubsection{Analysis on Association-Centric Learning Mechanism}
We first verified the effectiveness of the association-centric learning mechanism on TAO in Tab.~\ref{tab:asscociation-centric_ablation}.
We conduct extensive combinations of the three proposed modules in association-centric learning.
When using spatial contrastive learning ($\mathcal{L}_{\mathrm{spatial}}$) alone, the AssocA shows a significantly low value, i.e., 6.4\%.
This is because spatial contrastive learning only endows AED with a sense of spatial perception.
Temporal contrastive learning ($\mathcal{L}_{\mathrm{temporal}}$) aligns with the association process because $S_{\mathrm{temporal}}$ is provided for data association.
It can be seen from Tab.~\ref{tab:asscociation-centric_ablation} that simply using $\mathcal{L}_{\mathrm{temporal}}$ achieves a comparatively high TETA and AssocA.
Due to an excessive focus on global consistency, utilizing only cross-clip consistency ($\mathcal{L}_{\mathrm{cross-clip}}$) results in a suboptimal performance.
With the combination of $\mathcal{L}_{\mathrm{spatial}}$, $\mathcal{L}_{\mathrm{temporal}}$, and $\mathcal{L}_{\mathrm{cross-clip}}$, AED achieves the best results.

\subsubsection{Analysis on Sampling Strategy}
As the input of AED is a video clip in the training stage, there are different ways to merge a number of video frames into a clip.
We conduct ablation experiments on two sampling methods on the TAO dataset as shown in Tab.~\ref{tab:sampling_ablation}.
The first setting is that we fix the sampling interval to the number of three, i.e., skip two frames when sampling.
Second, we allowed the sampling interval a to change randomly within the range of [1, 3], which introduces a wider variety of displacement and perspective changes during training.
We find that the random sampling strategy improves the AssocA by 0.6 points.

\vspace{-0.5cm}

\begin{table}[t]
    \caption{Different sampling strategy ablation on the validation set of TAO dataset.}
    \label{tab:sampling_ablation}
    \centering
        \begin{tabularx}{8cm}
        {>{\centering\arraybackslash}p{3cm}|*{4}{>{\centering\arraybackslash}X}}
        \toprule
        Sampling Methods    & TETA          & LocA          & AssocA        & ClsA          \\
        \midrule
        fixed               & 54.9          & 71.7          & 51.5          & 41.6          \\ %
        random              & \textbf{55.2} & \textbf{71.8} & \textbf{52.1} & \textbf{41.7} \\ %
        \bottomrule
        \end{tabularx}
        \vspace{-0.2cm}
\end{table}

\begin{figure*}[!t]
  \centering
  \includegraphics[width=16cm]{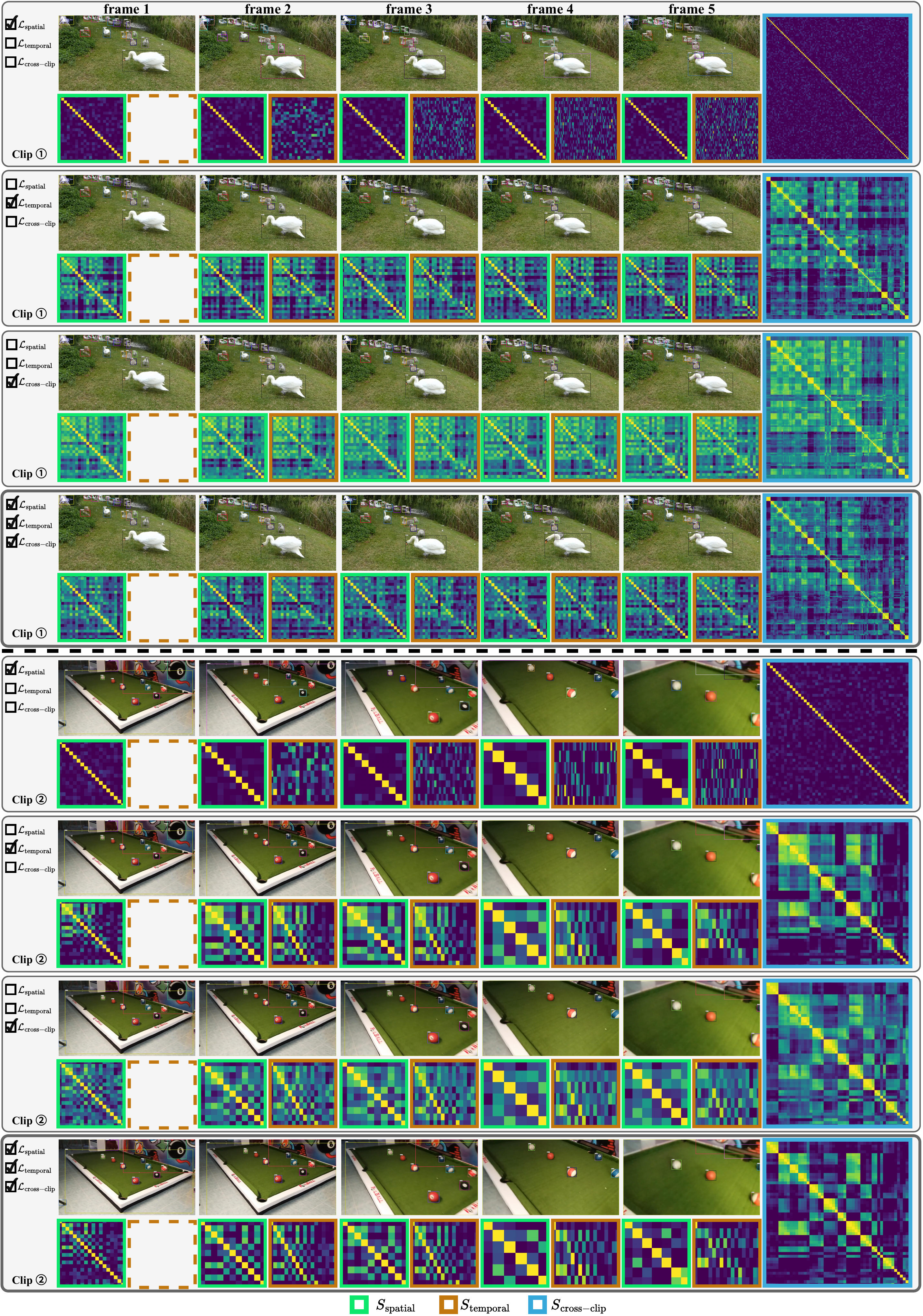}
  \caption{Visualize the effectiveness of association-centric learning. We show spatial similarities $S_{\mathrm{spatial}}$, temporal similarities $S_{\mathrm{temporal}}$, and cross-clip similarities $S_{\mathrm{cross-clip}}$ under different association-centric learning settings for clip \ding{172} and clip \ding{173}.
  The higher the value of each element in the $S$ matrix, the closer its color tends to be yellow.
  Conversely, the closer it tends to be blue.
  }
  \label{fig:vis_weight}
\end{figure*}

\begin{figure*}[!t]
  \centering
  \includegraphics[width=\textwidth]{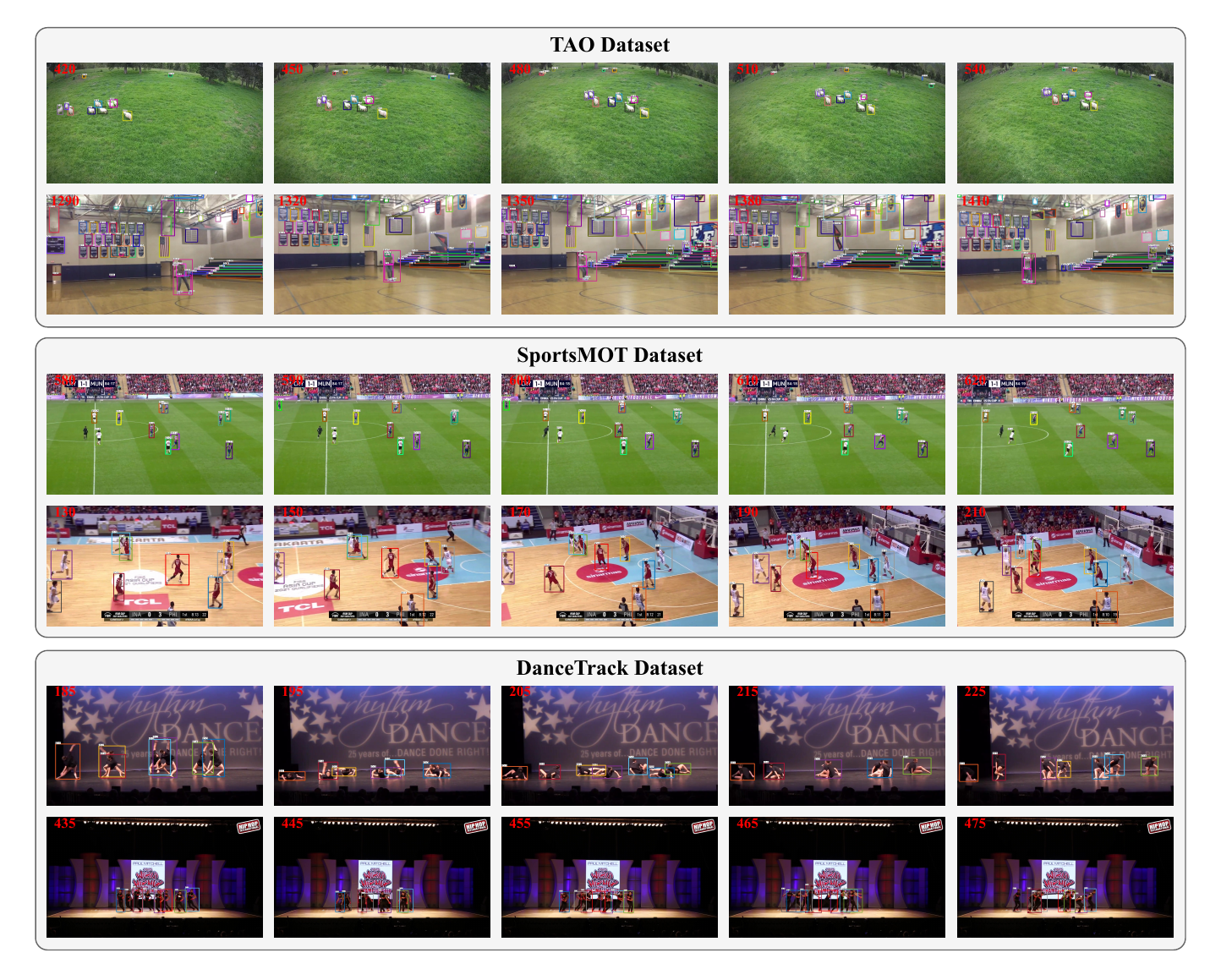}
  \caption{Successful and failed cases of AED. We visualize some tracking results from TAO, SportsMOT, and DanceTrack.
  }
  \label{fig:vis_tracking}
  \vspace{-0.5cm}
\end{figure*}

{\subsection{Analysis on Parameter Sensitivity and Performance Ceiling}

\begin{table}[t]
    \caption{Multi-head weight attention analysis on the validation set of TAO dataset.}
    \label{tab:weight_attention_ablation}
    \centering
        \begin{tabularx}{8cm}
        {>{\centering\arraybackslash}X|*{4}{>{\centering\arraybackslash}X}}
        \toprule
        $h$ & TETA          & LocA          & AssocA        & ClsA          \\
        \midrule
        1   & 55.1          & \textbf{71.8} & 51.8          & \textbf{41.7} \\ %
        2   & \textbf{55.2} & \textbf{71.8} & \textbf{52.1} & \textbf{41.7} \\ %
        4   & 54.8          & \textbf{71.8} & 51.0          & \textbf{41.7} \\ %
        8   & 55.0          & \textbf{71.8} & 51.6          & 41.6          \\ %
        16  & 54.6          & 71.7          & 50.5          & \textbf{41.7}  \\ %
        \bottomrule
        \end{tabularx}
        \vspace{-0.2cm}
\end{table}

\subsubsection{Weight Attention Analysis}
In Tab.~\ref{tab:weight_attention_ablation}, we conduct several experiments to validate the impact of different $h$ in the multi-head weight attention.
We select several different values of $h$, i.e., 1, 2, 4, 8, and 16.
It can be observed from Tab.~\ref{tab:weight_attention_ablation} that when $h$ equals 2, the model achieves a comparatively high performance.

\subsubsection{Analysis on IoU Threshold to Filter Out Ambiguous Queries}
As introduced in Section~\ref{sec:training}, after the refinement in training, we calculate IoUs between boxes of object queries and their GTs to filter out the ambiguous ones, i.e., IoU $< \tau_{\mathrm{iou}}$.
In order to investigate the impact of $\tau_{\mathrm{iou}}$ on AED, we conduct a comparative experiment in Tab.~\ref{tab:IoU_ablation}.
We find that a higher $\tau_{\mathrm{iou}}$ has a slight negative impact on the association performance (AssocA) of AED.
Ultimately, we chose $\tau_{\mathrm{iou}}=0.5$ during training.

\subsubsection{Analysis on the Length of Clips}
During training, we take sampled clips of $n$ frames as the input.
We conduct an experiment on the TAO dataset to find the most suitable $n$ in Tab.~\ref{tab:clip_length_ablation}.
We can find that with the increment of $n$, the performance of AED also improves.
However, due to the limitation of GPU memory, $n=5$ represents the current upper limit of our experimental conditions.

\subsubsection{Analysis on Performance Ceiling}
To further exhibit the upper limits of AED, we replace the detection boxes with GTs.
We compare AED with several other high-performance methods including ByteTrack~\cite{zhang2022bytetrack}, DeepSORT~\cite{wojke2017simple}, and OVTrack~\cite{li2023ovtrack} in Tab.~\ref{tab:ceiling_ablation}.
AED achieves the highest performance among TETA, AssocA, and ClsA metrics.

\begin{table}[t]
    \caption{$\tau_{iou}$ analysis on the validation set of TAO dataset.}
    \label{tab:IoU_ablation}
    \centering
        \begin{tabularx}{8cm}
        {>{\centering\arraybackslash}p{1.5cm}|*{4}{>{\centering\arraybackslash}X}}
        \toprule
        $\tau_{iou}$ & TETA          & LocA          & AssocA        & ClsA          \\
        \midrule
        0.1          & 55.1          & 71.8          & 51.8          & 41.7          \\ %
        0.3          & \textbf{55.2} & \textbf{71.9} & 51.9          & 41.7          \\ %
        0.5          & \textbf{55.2} & 71.8          & \textbf{52.1} & 41.7          \\ %
        0.7          & 55.0          & 71.8          & 51.5          & 41.7          \\ %
        0.9          & 54.3          & 71.7          & 49.5          & \textbf{41.8} \\ %
        \bottomrule
        \end{tabularx}
        \vspace{-0.2cm}
\end{table}

\subsection{Visualization}

\subsubsection{Association-Centric Learning}
In Fig.~\ref{fig:vis_weight}, we visualize the three proposed similarity matrices, i.e., $S_{\mathrm{spatial}}$, $S_{\mathrm{temporal}}$, and $S_{\mathrm{cross-clip}}$, from two scenes under different settings to validate the effectiveness of association-centric learning.
We arrange the rows and columns of each $S$ in the order of the IDs.
Ideally, high-response elements (yellow ones) should be along the diagonal or within the diagonal blocks.
For each clip, we also visualize its tracking result and represent different IDs with bounding boxes of different colors.
When only adopting spatial contrastive learning ($\mathcal{L}_{\mathrm{spatial}}$), the model identifies every detected object as a unique ID due to the lack of temporal supervision (first row of clip \ding{172} and \ding{173}).
As mentioned above, $S_{\mathrm{cross-clip}}$ represents similarities between every collected object query in the entire clip.
This can be further verified from another perspective that $S_{\mathrm{cross-clip}}$ has no high-response block along the diagonal.
Simply using temporal contrastive learning ($\mathcal{L}_{\mathrm{temporal}}$) achieves moderately good results (second row of clip \ding{172} and \ding{173}).
However, compared with the last row of each clip, some erroneous high-response regions can still be observed especially in the $S_{\mathrm{cross-clip}}$ of clip \ding{173}.
When using cross-clip contrastive learning ($\mathcal{L}_{\mathrm{cross-clip}}$) alone, all of the elements of $S_{\mathrm{spatial}}$, $S_{\mathrm{temporal}}$, and $S_{\mathrm{cross-clip}}$ in clip \ding{172} and clip \ding{173} have relatively high responses (third row of clip \ding{172} and \ding{173}).
This is because of the overemphasis on global feature consistency.
When making full use of association-centric learning, the discriminative ability of AED is further enhanced in the last row of clip \ding{172} and \ding{173}.
We can see from $S_{\mathrm{spatial}}$ that AED can distinguish between each object within a single frame very well, as almost every $S_{\mathrm{spatial}}$ matrix exhibits high responses only along the diagonal.
Also, it can be observed from the $S_{\mathrm{temporal}}$ matrices that nearly every track query is matched correctly even with very similar appearances in clip \ding{172}.
It can be observed that almost all high-similarity blocks are exclusively found on the diagonal of $S_{\mathrm{cross-clip}}$.

\begin{table}[t]
    \caption{The impact of different clip lengths on the performance of AED.}
    \label{tab:clip_length_ablation}
    \centering
        \begin{tabularx}{8cm}
        {>{\centering\arraybackslash}p{1.5cm}|*{4}{>{\centering\arraybackslash}X}}
        \toprule
        Clip Length  & TETA        & LocA          & AssocA        & ClsA          \\
        \midrule
        2          & 53.6          & \textbf{71.8} & 47.3          & \textbf{41.7} \\ %
        3          & 54.9          & 71.7          & 51.3          & \textbf{41.7} \\ %
        4          & 55.1          & 71.7          & 51.8          & \textbf{41.7} \\ %
        5          & \textbf{55.2} & \textbf{71.8} & \textbf{52.1} & \textbf{41.7} \\ %
        \bottomrule
        \end{tabularx}
        \vspace{-0.2cm}
\end{table}

\begin{table}[t]
    \caption{To show the performance ceiling of AED, we compare several OV-MOT trackers with AED by using GT bounding boxes during inference.}
    \label{tab:ceiling_ablation}
    \centering
        \begin{tabularx}{8.5cm}
        {l|*{4}{>{\centering\arraybackslash}X}}
        \toprule
        \multicolumn{1}{c|}{Methods}        & TETA          & LocA          & AssocA        & ClsA          \\
        \midrule
        ByteTrack~\cite{zhang2022bytetrack} & 75.0          & 99.2          & 27.2          & 98.7          \\ %
        Deepsort~\cite{wojke2017simple}     & 75.1          & 96.3          & 33.2          & 95.9          \\ %
        OVTrack~\cite{li2023ovtrack}        & 87.6          & \textbf{99.5} & 71.9          & 91.5          \\ %
        AED (Ours)                          & \textbf{92.3} & 99.2          & \textbf{77.7} & \textbf{99.9} \\ %
        \bottomrule
        \end{tabularx}
        \vspace{-0.2cm}
\end{table}

\subsubsection{Tracking Results}
In Fig.~\ref{fig:vis_tracking}, we visualize some successful and failed tracking results of AED on different datasets including TAO, SportsMOT, and DanceTrack.
Specifically, we use Co-DETR in TAO and YOLOX in SportsMOT and DanceTrack.
AED is able to handle a wide range of scenarios including low frame rates (TAO), large movements(SportsMOT), severe occlusions (DanceTrack), and similar appearances (SportsMOT and DanceTrack).
However, AED makes some mistakes in cases involving tiny objects (ID10, 12, etc of the first row of the TAO dataset) and highly dense scenes (ID 64, 65, etc of the second row of the TAO dataset).

\section{Conclusion}
In this paper, we propose AED to apply to OV-MOT and CV-MOT tasks following the tracking-by-detection paradigm.
AED is compatible with various detectors and unknown categories.
We employ a very simple matching strategy by using the proposed sim-decoder without any prior knowledge.
Moreover, we propose association-centric learning to ensure that the training objective remains consistent with the needs of the association phase.
We also conducted sufficient experiments to validate the superior performance of our method compared with existing OV-MOT and CV-MOT trackers.

However, AED also has its limitations:
\begin{itemize}
    \item AED struggles to tackle extremely crowded scenarios due to the lack of motion cues.
    \item Object queries of AED are quite homogeneous, originating solely from the detector and lacking information from other forms or modalities, such as text.
\end{itemize}

\bibliographystyle{IEEEtran}
\bibliography{main}

\vfill

\end{document}